\documentclass{llncs}
\usepackage{url}
\usepackage{amsmath}
\usepackage{todonotes}
\title{Leipzig Corpus Miner -- A Text Mining Infrastructure for Qualitative Data Analysis}

\begin{document}

\author{Andreas Niekler
\and Gregor Wiedemann
\and Gerhard Heyer}

\institute{University of Leipzig \\
   Augustusplatz 10 \\
   04109 Leipzig, Germany \\
   \url{{aniekler, heyer}@informatik.uni-leipzig.de}, \url{gregor.wiedemann@uni-leipzig.de} }
   

\maketitle
\begin{abstract}
This paper presents the ``Leipzig Corpus Miner''---a technical infrastructure for 
supporting qualitative and quantitative content analysis. The infrastructure
aims at the integration of ``close reading'' procedures on individual documents
with procedures of ``distant reading'', e.g. lexical characteristics of large
document collections. Therefore information retrieval systems, lexicometric
statistics and machine learning procedures are combined in a coherent framework
which enables qualitative data analysts to make use of state-of-the-art Natural
Language Processing techniques on very large document collections. Applicability
of the framework ranges from social sciences to media studies and market
research. As an example we introduce the usage of the framework in a political
science study on post-democracy and neoliberalism.
\end{abstract}

\section{Introduction}

For several years humanists, social scientists and media analysts working with
text as primary data have been opening up to large scale text analysis
procedures. This development, sometimes referred to by the term ``Digital Humanities'', is
fueled by increased availability of digital text and algorithms
for identification of (semantic) structures in unstructured data. Nonetheless,
the complexity of handling `Text Mining' (TM) procedures as well as problems in
managing of `big data' prevents those approaches from being used by a wider
audience lacking a computer science background.

To facilitate the handling of large document collections and make use of
algorithmic Text Mining procedures for content analysis (CA) we built the
``Leipzig Corpus Miner'' (LCM). With CA we refer to a broad set of methods and
corresponding methodologies for analyzing textual data common in various
scientific disciplines. These may include classic (quantitative) content
analysis~\cite{Krippendorff.2013} as well as rather qualitative approaches like
discourse analysis~\cite{Laclau.2001}, grounded theory \cite{Glaser.2005} or
qualitative content analysis~\cite{Mayring.2010}. In this respect, techniques
integrated into the LCM do not replicate analysis procedures of these
methods exactly. Rather they offer a set of tools which enable analysts to support certain steps of an applied method and to extend the size of collections under investigation to a
degree which could not be handled manually.

In contrast to most computer-assisted CA studies which employ only a single or
very few TM procedures~\cite{Wiedemann.2013} LCM allows for
application of multiple procedures which may be integrated systematically into complex analysis workflows. Thus,
results of single processes are not restricted to be interpreted in an isolated
manner. Beyond that, they may be used as input data for further mining
processes. For example, a list of key terms automatically extracted from a
subset of a reference corpus may be utilized to retrieve documents of interest
in a second target corpus. Semantic topics automatically retrieved from
Topic Models~\cite{Blei.2012} may be applied to disambiguate homonymous term
usage and thus, help defining concepts under investigation for individual CA
studies.

The systematic integration of TM procedures in a coherent user-friendly
application enables content analysts to employ complex algorithms while
simultaneously retaining control over major parameters or configurations of
analysis process chains. Utilizing those procedures on large document
collections may improve quality of qualitative data analysis especially in terms
of reliability and reproducibility. It thereby gives analysts without deeper
knowledge of foundations of natural language processing (NLP) the ability to
develop best practices for computer-assisted large scale text analysis. Because NLP experts usually lack background knowledge about requirements and methodologies in the
humanities and social sciences such best practices could not be
developed by either discipline alone. Providing a common interface on text analysis is
the major advantage of using a framework like the LCM.

\section{Architecture}

\begin{figure}[t]
\centering
\includegraphics[width=8.5cm]{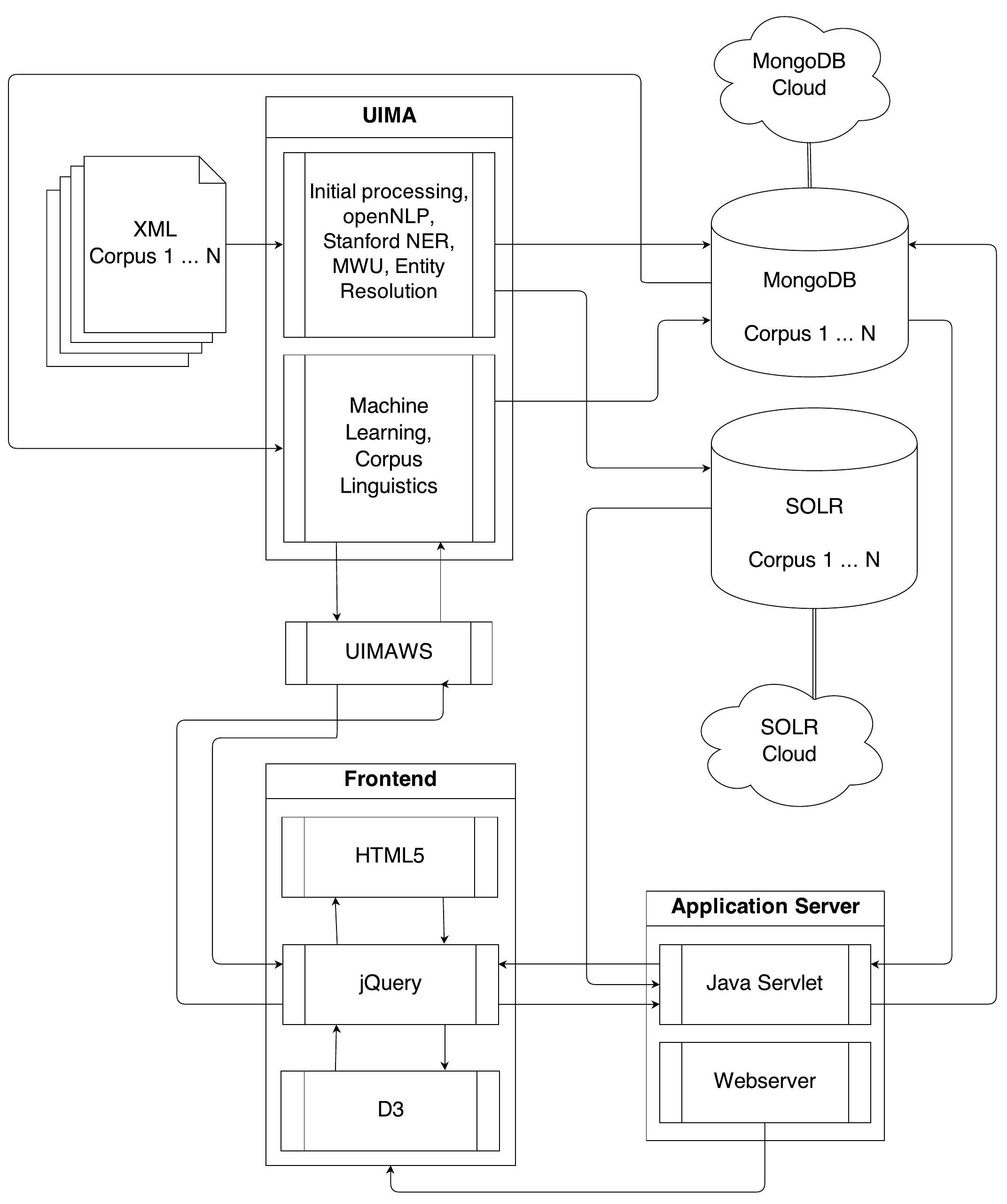}
\caption{Architecture of the LCM.\label{fig:arch}}
\end{figure}

The LCM is a combination of different technologies which provide a
qualitative data analysis environment accessible by an interface targeted towards domain experts unfamiliar with NLP.
Analysts are put in a position to work on their data with more methodical rather than technical understanding of the algorithms. Applied technologies behind the user interface need to support analysts in tasks such as \emph{data storage}, \emph{retrieval}, \emph{processing} and \emph{presentation}. A schematic overview of the architecture and workflow is given in figure~\ref{fig:arch}.

\par Storage and processing of large amounts of text data are key tasks
within the proposed environment. To process data we use UIMA, an
architecture to identify structures in unstructured textual data~\cite{Ferrucci.2004}. Within this architecture data readers, data writers and data processing classes can be chained together in order to define standardized workflows for datasets. 
In UIMA data processing is done by so called ``annotators'' which (usually) add additional information to the analyzed documents. UIMA employs ``stand off'' markup for annotations which in contrast to inline markup does not alter the original text. Therefore, annotations can be stored in a separate data structure~\cite{burghardt_stand_2009}.
Annotations could include sentence information, token separations or
classification labels. But annotators can also be used to aggregate single text documents for further processing such as topic modeling. We defined UIMA processes for the following
tasks:
\begin{itemize}
  \item \textbf{Initial processing of XML sources:} To initially process raw
  sets of text data we defined a process which identifies areas like titles and
  paragraphs, separates sentences, applies tokenization and POS-Tagging, detects
  entities and multi word units (MWU) and performs entity resolution. 
  \textit{OpenNLP}\footnote{http://opennlp.apache.org}, the \textit{Stanford NER}
  tool~\cite{finkel_incorporating_2005}, a Wikipedia based multi word unit
  detection and the \textit{JRC-Names} resource~\cite{steinberger_jrc-names.2013} for entity resolution are utilized within
  this tool chain. We trained custom models for sentence detection
  and tokenization with resources provided by the \textit{Leipzig Corpora Collection}
  (LCC)~\cite{BHQ+07}.
  \item \textbf{Classification, Corpus Linguistics (CL) and Machine Learning
  (ML):} The UIMA based \textit{ClearTK}~\cite{ogren-etal.2008} framework is applied to provide convenient access to
  classification and machine learning libraries to support certain analysis steps. We also implemented
  inference algorithms for different Topic Models and corpus linguistic analyses
  based on ClearTK.
\end{itemize}

\noindent Annotations made by processing chains, e.g. tokens, sentences,
entities or classification labels, can vary throughout different analysis
approaches. We therefore decided to employ a schema free data storage which gives us the flexibility to delete or add annotation structures to stored text
documents. For this, we use MongoDB\footnote{http://www.mongodb.org}---a
NoSQL database implementation which stores datasets as JavaScript Object Notation (JSON) structures. It
supports the distribution of very large datasets to different shards and
machines. This is
a very important feature as we want to open the system for new large corpora. Every imported corpus can be accessed separately within the architecture.
\par Alongside with data storage the user-friendly accessibility of corpora by
different indexes is an additional requirement. As we want to provide indexes accessible by a
convenient query language based on metadata fields and full-text of documents we use the \textit{Solr}\footnote{http://lucene.apache.org/solr} enterprise search server. For this a set of metadata fields and full-text contents for indexing each corpus can be defined. This information is provided to Solr which uses 
annotations made by UIMA directly to populate its indexes. As well as MongoDB we
can distribute the Solr indexes to a cloud and thus are able to index and store new
large corpora. Indexes are used for full-text search, faceted search and search on meta data fields. 

\begin{figure}
\begin{minipage}[hbt]{5.8cm}
\centering
\includegraphics[width=5.8cm]{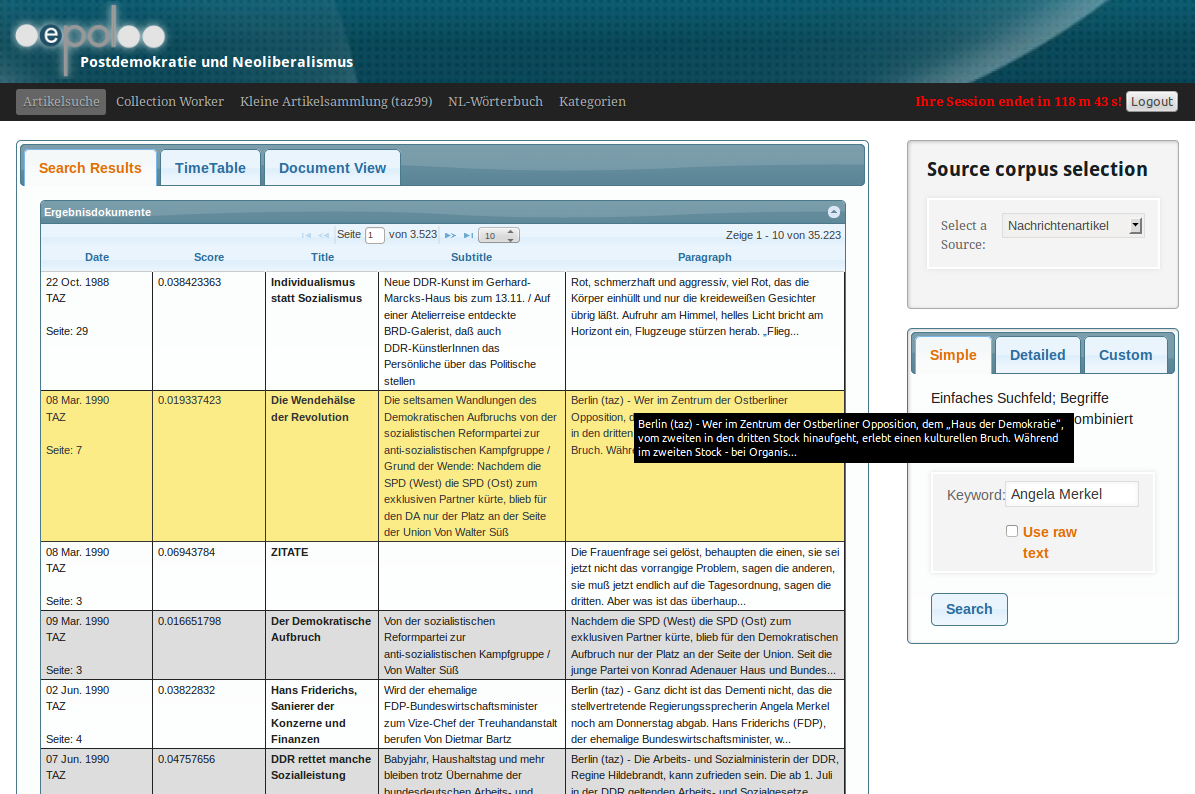}
\caption{Browser-based user interface of the LCM.\label{fig:lcm}}
\end{minipage}
\hfill
\begin{minipage}[hbt]{5.8cm}
\centering
\includegraphics[width=5.8cm]{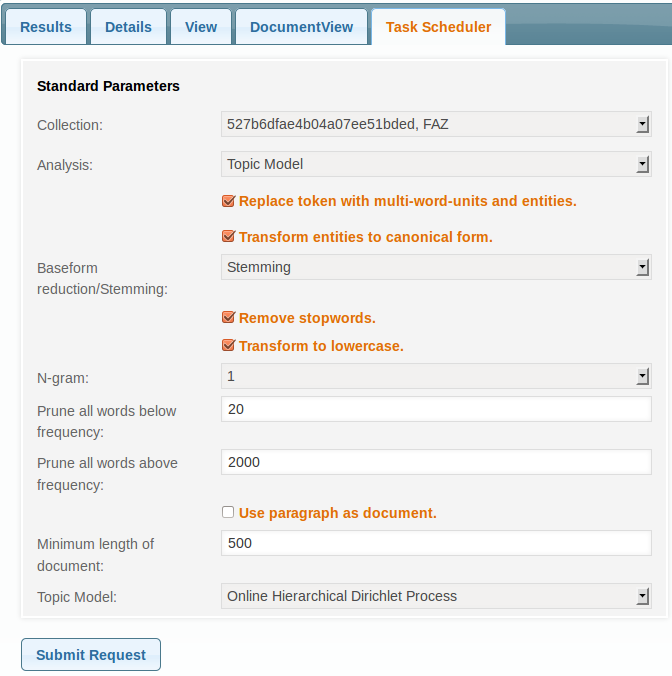}
\caption{The GUI enables users to manage various Text Mining tasks on sub
collections.\label{fig:task}}
\end{minipage}
\end{figure}

Access to
the datasets and indexes is facilitated by a web-based front end (fig.~\ref{fig:lcm}). This web application, based on the \textit{YAML} CSS framework\footnote{http://www.yaml.de}, accesses stored corpora and
datasets through Java Servlets\footnote{http://www.oracle.com/technetwork/java/index-jsp-135475.html} hosted by an Application Server. The front end to back end communication is managed by \textit{AJAX} calls via the \textit{jQuery}\footnote{http://jquery.com} javascript framework. Presentations of results, graphs and charts are implemented in \textit{D3}\cite{bostock.2011}, a visualization library to create, manipulate and animate SVG Objects.

\par One central objective of the LCM is to enable analysts to perform Text Mining tasks without explicit guidance by NLP experts. For that reason we implemented a middleware (UIMAWS), a
webservice to start, stop and manage UIMA processes for certain tasks. The
analysis tasks are described in detail in section~\ref{sec:capab} of this
paper. The UIMAWS is deployed to a dedicated server and is able to manage
multiple instances of UIMA processes. The webservice communicates information on the progress of running processes. Results of completed tasks will be written to the database and visualized by the front end. The UIMA processes always operate on
a finite set of documents which are referenced by their ID within
MongoDB. Those identifications can be retrieved by querying the Solr index, a custom information retrieval algorithm or manually compiled lists of documents. Starting and managing of UIMA processes can be done in the front end as shown in figure~\ref{fig:task}.

\section{Analysis capabilities}
\label{sec:capab}

The LCM integrates several procedures for retrieving, annotating and mining
textual data. Flexibility in combining these tools lends support to various
analysis interests ranging from quantitative corpus linguistics to qualitative
reconstructivist methodologies. We briefly introduce the inbuilt technologies
and give an example of a workflow in a concrete research environment in section
\ref{sec:example}.

\begin{figure}
\begin{minipage}[hbt]{12.2cm}
\centering

\includegraphics[width=9cm]{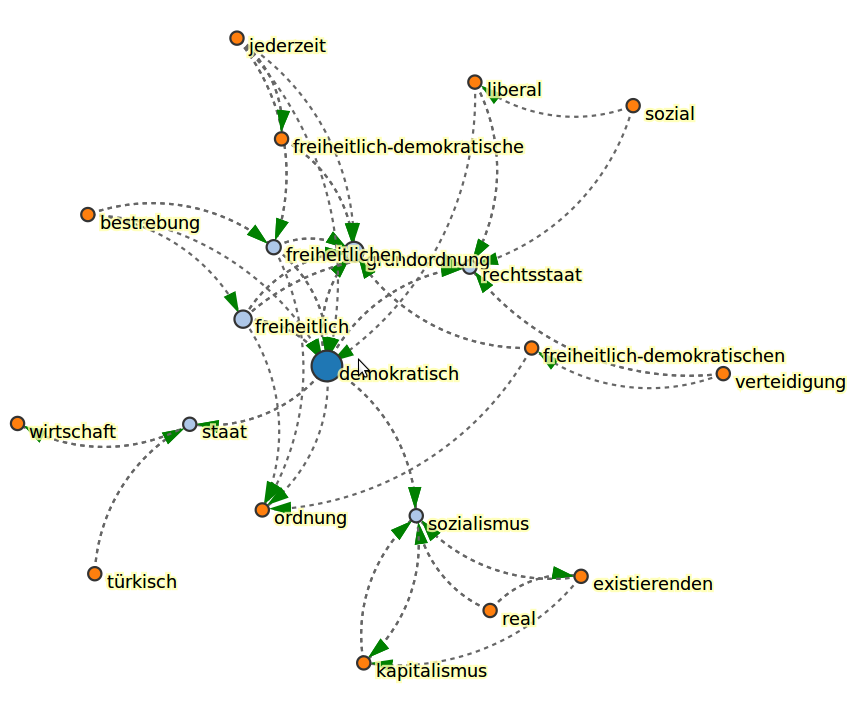}
\caption{The LCM provides various data visualizations on sub collections, e.g.
co-occurrence graphs...\label{fig:coocs}}

\hfill

\includegraphics[width=9cm]{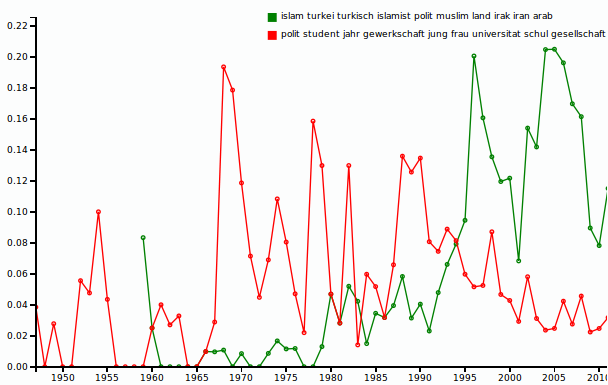}
\caption{... or topic model distributions over time.\label{fig:tms}}
\hfill

\includegraphics[width=9cm]{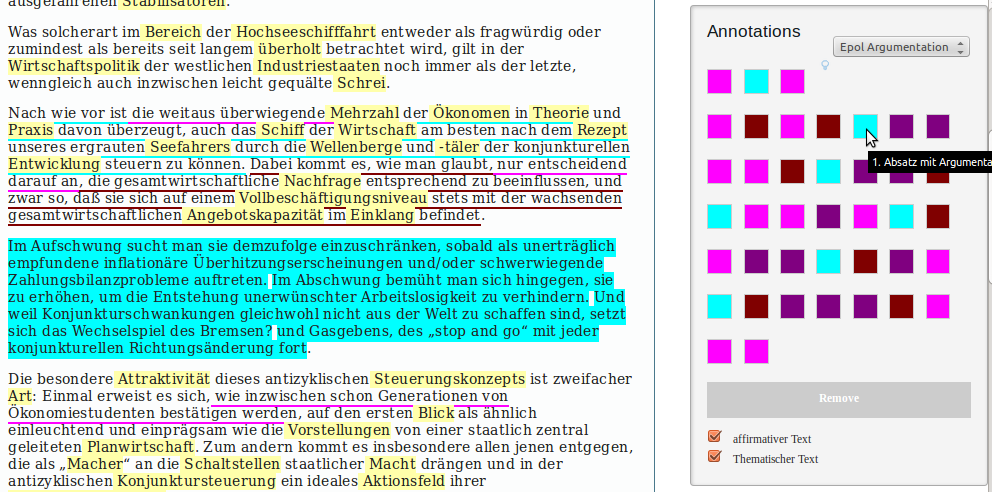}
\caption{The UI provides assistance for manual annotation of document
parts which subsequently may be used as training examples for classification.\label{fig:annot}}
\end{minipage}
\end{figure}

\begin{itemize}
\item \textbf{Information retrieval:} Assuming the availability of a large
document collection, e.g. complete volumes of a daily newspaper over several
decades, a common need is to identify documents of interest for certain
research questions. The LCM provides two ways of IR for this task. First, a full
text index allows for key term search over the entire collection (see section
\ref{fig:lcm}). But, for CA relevancy often cannot be tied to a handful of
key terms. Thus, we provide a second retrieval method based on contextualized
dictionaries which can be retrieved from a reference corpus of paradigmatic
documents covering a certain research interest. A list of several hundred key terms, called `dictionary', is extracted
automatically from this reference corpus and co-occurrence patterns of these
terms are measured. Relevancy in our target collection is then defined through
occurrence of dictionary terms in their desired co-occurrence contexts from the reference
collection. This method allows not only for retrieving coherent document sets on
a single subject-matter (e.g. `Iraq war'), but rather for distinguishing modes of
speech over various topics (e.g. military cadence in politics or sports).
Document collections retrieved by either IR system can be stored for further
procedures.
\item \textbf{Lexicometrics:} The LCM has implemented computation and
visualization of basic corpus linguistic measures on stored collections. It
allows for frequency analysis, co-occurrence analysis (figure \ref{fig:coocs}) and
automatic extraction of key terms. As documents contain a time-stamp in their
metadata, visualizations can aggregate these measurements over days, weeks, months or
years. Researchers may choose if they want to have absolute or relative
frequencies displayed. For co-occurrence analysis LCM allows selecting
different significance measures to compute term relations (raw counts, Tanimoto,
Dice, Mutual-Information, Log-Likelihood)\cite{Buchler.2008}.
\item \textbf{Topic models}: For analysis of topical structures in large text
collections Topic Models have been shown to be useful in recent studies
\cite{Niekler.2012}. Topic Models are statistical models which infer probability
distributions over latent variables, assumed to represent topics, in text collections as
well as in single documents. So far the LCM has implemented models of Latent
Dirichlet Allocation (LDA)~\cite{Blei.2003}, Hierarchical Dirichlet Process
(HDP), Pitman-Yor Process (HPY)~\cite{TehJor2010a} and Online LDA~\cite{hoffman_online_2010}, a very fast inference algorithm for the LDA model~\cite{Blei.2003}.
Results from Topic Models can be seen as an unsupervised clustering which gives analysts the opportunity to identify and
describe thematic structures on a macro level as well as to select subsets of
documents from a collection for further refinement of the analysis or for a
``close reading'' process. Distributions of topics can be visualized over time
(figure \ref{fig:tms}).
\item \textbf{Classification:} Supervised learning from annotated text to assist
coding of documents or parts of documents promises to be one major
innovation to CA applications~\cite{Scharkow.2012}. The LCM allows for manual
annotation of complete documents or snippets of documents with category labels
(figure \ref{fig:annot}). The analyst may initially develop a hierarchical category
system and / or refine it during the process of annotation. Annotated text parts
are used as training examples for automatic classification processes which
output category labels for unseen analysis units (e.g. sentences, paragraphs or
documents). For that the LCM integrated several classification approaches like
SVM, Naive Bayes and Semantically Smoothed Naive Bayes~\cite{Zhou.2008}. Feature
engineering can be performed by the analyst up to a certain degree by
configuring the classification process (e.g. restriction to certain POS-types
for training). An iterated process of manual labeling and evaluation of (best)
automatically retrieved labels may replicate ``Hybrid Active Learning''~\cite{Lughofer.2012}.
\end{itemize}

\par

\section{Example use case}
\label{sec:example}

\par The LCM was initially built to facilitate a CA study on political theory.
Within the German ``ePol-project''\footnote{\url{http://www.epol-project.de}}
political scientists conduct large scale text analyses with the help of NLP
researchers. The project aims to identify changes in discursive
patterns of policy justifications in public media. By identifying certain
patterns and measuring their quantities the project strives to verify or reject
central hypotheses about a phenomenon called ``Post-democracy''. One central
question is: Has there been an ``economization'' of justifications in some or all policy fields during the last decades?~\cite{Lemke.2012}. To answer this
question a corpus consisting of 3.5 million newspaper articles from 1946 to 2012 is
investigated.\footnote{The corpus consists of complete volumes of \textit{DIE ZEIT, TAZ, S\"uddeutsche Zeitung} and a representative sample of \textit{Frankfurter Allgemeine Zeitung.}}

\par
\begin{figure}
\centering
\includegraphics[width=12.2cm]{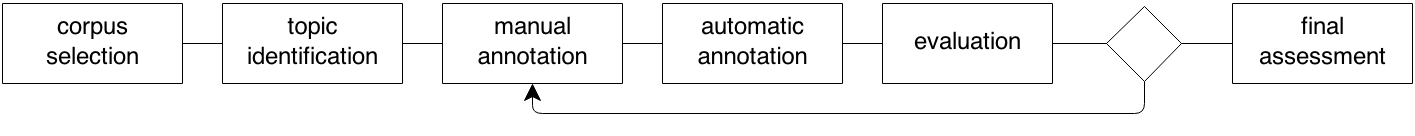}
\caption{Schema of a Content Analysis workflow realized within the
LCM.\label{fig:workflow}}
\end{figure}

\par The LCM was built with the requirement to support analyses of and inquiries into these rather
abstract questions. Figure \ref{fig:workflow} shows a simplified workflow of the
analysis process which is realized with the help of the aforementioned
procedures.

\par
\begin{enumerate}
\item A subset of relevant documents for the analysis is retrieved from the
complete corpus of 3.5 million documents. To identify ``neoliberal'' themes and
modes of speech a reference corpus consisting of 36 works from neoliberal theorists
was compiled. A dictionary of 500 automatically extracted key terms and their sentence
co-occurrences were computed. With our IR system we select 10,000 (potentially) relevant documents from the initial corpus.
\item A topic model computation over this subset identifies thematic clusters
which help to describe its content from a macro perspective. The topic model
result allows to distinguish of policy fields mentioned within the articles. It also enables analysts to remove documents from the selected subset which contain topics irrelevant to the specific purpose of this analysis (e.g. articles mainly concerned with foreign policy are not considered relevant for this study).
\item Best ranked documents from step 1 of each topic (step 2) are investigated
manually by the political scientists to identify argumentative structures in
different policy fields. A hierarchical category system is derived from common, distinctive structures. Relevant sections of the documents were then annotated with category labels.
\item An automatic classification procedure is invoked on the unlabeled data to
identify more text parts containing argumentative structures. The NLP group
supports the analysts by identifying discriminating linguistic features for this task.
\item Text snippets identified in the previous step (supposedly containing
arguments of interest) are presented to the analysts ranked by
certainty of label assignments. Analysts can verify or reject the results
manually. In this active learning paradigm we calculate internal precision /
recall measures while the analysts are evaluating the process qualitatively. If those 
ongoing evaluations show satisfactory results, the process of creating training data is concluded.
\item The classification procedure is run on the entire collection under
investigation. Results can be described qualitatively and quantitatively (e.g.
proportions of categories over time slices).
\end{enumerate}

\section{Future Work}

\par The LCM supports manual Content Analysis (CA) via basic corpus linguistic
procedures as well as supervised state-of-the-art Text Mining techniques. In a next step we will incorporate a user rights management system which allows for access restrictions to
different corpora in the LCM. In addition, work-flows will be further enhanced to combine intermediate results of different procedures. For example, a centralized management of
(semi\-)automatically generated dictionaries could be useful to exploit
controlled lists of key terms throughout different steps of the entire process
chain (e.g. computing co-occurrences just for key terms; using occurrence of terms
in dictionaries as additional feature for classification tasks). To support more inductive research approaches, we plan to integrate unsupervised data-driven analysis procedures into the LCM. These could help analysts especially between steps 2 and 3 of our proposed process chain to identify categories of interest in unknown data. This
includes the identification of stable as opposed to volatile semantic concepts in
certain topics over time~\cite{Heyer.2009c}. 

In classification and annotation
processes we enable analysts to define categories and training data
for those categories. Within this process many evaluation measures, frequently used by CA scholars, could be applied. In future developments we will incorporate more of
these measures regarding cooperative annotation and categorization tasks. This
will be a further step towards integration with common social science research methods away from a narrow computer linguistic oriented work-flow. Its our hope that the LCM may contribute to improve the comparability, reliability and validity of social science research standards.

\bibliographystyle{splncs03}
\bibliography{bibdata}

\begin{thebibliography}{10}
\providecommand{\url}[1]{\texttt{#1}}
\providecommand{\urlprefix}{URL }

\bibitem{BHQ+07}
Biemann, C., Heyer, G., Quasthoff, U., Richter, M.: The leipzig corpora
  collection - monolingual corpora of standard size. In: Proceedings of Corpus
  Linguistics Conference (2007)

\bibitem{Blei.2012}
Blei, D.M.: Probabilistic topic models: Surveying a suite of algorithms that
  offer a solution to managing large document archives. Communications of the
  ACM  55(4),  77--84 (2012)

\bibitem{Blei.2003}
Blei, D.M., Ng, A.Y., Jordan, M.I.: Latent dirichlet allocation. Journal of
  Machine Learning Research  3,  993--1022 (2003)

\bibitem{bostock.2011}
Bostock, M., Ogievetsky, V., Heer, J.: D3: Data-driven documents. IEEE Trans.
  Visualization \& Comp. Graphics (Proc. InfoVis)  17(4),  2301 -- 2309 (2011)

\bibitem{Buchler.2008}
B{\"u}chler, M.: Medusa: Performante Textstatistiken auf gro{\ss}en Textmengen:
  Kookkurrenzanalyse in Theorie und Anwendung. VDM (2008)

\bibitem{burghardt_stand_2009}
Burghardt, M., Wolff, C.: {Stand off-Annotation f{\"u}r Textdokumente: Vom
  Konzept zur Implementierung (zur Standardisierung?)}. In: Proceedings of the
  Biennial GSCL Conference. pp. 53--59 (2009)

\bibitem{Ferrucci.2004}
Ferrucci, D., Lally, A.: Uima: An architectural approach to unstructured
  information processing in the corporate research environment. Nat. Lang. Eng.
   10(3-4),  327--348 (2004)

\bibitem{finkel_incorporating_2005}
Finkel, J.R., Grenager, T., Manning, C.: Incorporating non-local information
  into information extraction systems by gibbs sampling. In: Proceedings of the
  43rd Annual Meeting on Association for Computational Linguistics. pp.
  363--370 (2005)

\bibitem{Glaser.2005}
Glaser, B.G., Strauss, A.L.: Grounded theory: Strategien qualitativer
  Forschung. Huber, 2 edn. (2005)

\bibitem{Heyer.2009c}
Heyer, G., Holz, F., Teresniak, S.: Change of topics over time and tracking
  topics by their change of meaning. In: Proceedings of the International
  Conference on Knowledge Discovery and Information Retrieval. pp. 223--228
  (2009)

\bibitem{hoffman_online_2010}
Hoffman, M.D., Blei, D.M., Bach, F.R.: Online learning for latent dirichlet
  allocation. In: Neural Information Processing Systems (NIPS). pp. 856--864
  (2010)

\bibitem{Krippendorff.2013}
Krippendorff, K.: Content analysis: An introduction to its methodology. SAGE, 3
  edn. (2013)

\bibitem{Laclau.2001}
Laclau, E., Mouffe, C.: Hegemony and socialist strategy. Verso, 2 edn. (2001)

\bibitem{Lemke.2012}
Lemke, M.: {Die {\"O}konomisierung des Politischen: Entdifferenzierungen in
  kollektiven Entscheidungsprozessen. Discussion Paper Nr. 2}. Hamburg and
  Leipzig (2012),
  \url{http://www.epol-projekt.de/discussion-paper/discussion-paper-2/}

\bibitem{Lughofer.2012}
Lughofer, E.: Hybrid active learning (hal) for reducing the annotation efforts
  of operators in classification systems. Pattern Recognition  45(2),  884--896
  (2012)

\bibitem{Mayring.2010}
Mayring, P.: Qualitative Inhaltsanalyse: Grundlagen und Techniken. Beltz, 11
  edn. (2010)

\bibitem{Niekler.2012}
Niekler, A., J{\"a}hnichen, P., Heyer, G.: {ASV Monitor: Creating}
  comparability of machine learning methods for content analysis. In:
  Proceedings of the 2012 European Conference on Machine Learning and Knowledge
  Discovery in Databases - Volume Part II. pp. 812--815 (2012)

\bibitem{ogren-etal.2008}
Ogren, P.V., Wetzler, P.G., Bethard, S.: {ClearTK}: A {UIMA} toolkit for
  statistical natural language processing. In: Towards Enhanced
  Interoperability for Large {HLT} Systems: {UIMA} for {NLP} workshop at
  Language Resources and Evaluation Conference ({LREC}). pp. 865--869 (2008)

\bibitem{Scharkow.2012}
Scharkow, M.: Automatische Inhaltsanalyse und maschinelles Lernen. epubli, 1
  edn. (2012)

\bibitem{steinberger_jrc-names.2013}
Steinberger, R., Pouliquen, B., Kabadjov, M., Belyaeva, J., van~der Goot, E.:
  {JRC-NAMES: A} freely available, highly multilingual named entity resource.
  In: Proceedings of the International Conference Recent Advances in Natural
  Language Processing. pp. 104--110 (2011)

\bibitem{TehJor2010a}
Teh, Y.W., Jordan, M.I.: Hierarchical {B}ayesian nonparametric models with
  applications. In: Bayesian Nonparametrics. 1 edn. (2010)

\bibitem{Wiedemann.2013}
Wiedemann, G.: Opening up to big data: Computer-assisted analysis of textual
  data in social sciences. Forum Qualitative Sozialforschung / Forum:
  Qualitative Social Research  14(2) (2013)

\bibitem{Zhou.2008}
Zhou, X., Zhang, X., Hu, X.: Semantic smoothing for bayesian text
  classification with small training data. In: Proceedings of the SIAM
  International Conference on Data Mining. pp. 289--300 (2008)

\end{thebibliography}

\end{document}